\documentclass[runningheads]{llncs}

\usepackage{graphicx}
\usepackage{hyperref}
\usepackage{textcomp}
\usepackage{bbding}
\usepackage{amsmath}
\usepackage{amssymb}
\usepackage{mathtools}
\usepackage[font=footnotesize,labelfont=bf]{subcaption}
\usepackage{booktabs}
\usepackage{multirow}
\usepackage{pgf}
\usepackage{tikz}
\usetikzlibrary{arrows, automata, shapes, petri, positioning, calc}
\usepackage{pgfplots}
\usepackage[export]{adjustbox}

\hypersetup{
	colorlinks=true,
	linkcolor={red!50!black},
	citecolor={blue!60!black},
	urlcolor={blue!80!black},
	pdfauthor={Pegoraro Marco},
	pdftitle={{Probabilistic and Non-Deterministic Event Data in Process Mining: Embedding Uncertainty in Process Analysis Techniques}},
	pdfsubject={Process Mining over Uncertain Data},
	pdfkeywords={Process Mining, Process Science, Event Data, Probabilistic Data, Non-Deterministic Data},
	pdfproducer={LaTeX},
	pdfcreator={pdfLaTeX},
	bookmarksopen=true
}

\def\nomove{\gg}

\begin{document}

\title{Probabilistic and Non-Deterministic Event Data in Process Mining: Embedding Uncertainty\\in Process Analysis Techniques\thanks{I am very grateful to Prof. Wil van der Aalst, who advises my doctoral studies, and to Dr. Merih Seran Uysal, who supervises me in researching this topic. I thank the Alexander von Humboldt (AvH) Stiftung for supporting my research interactions.}}

\author{Marco Pegoraro\orcidID{0000-0002-8997-7517}}

\authorrunning{Marco Pegoraro}
\titlerunning{Embedding Uncertainty in Process Analysis Techniques}

\institute{Chair of Process and Data Science (PADS) \\ Department of Computer Science, RWTH Aachen University, Aachen, Germany
	\email{pegoraro@pads.rwth-aachen.de}\\
	\url{http://www.pads.rwth-aachen.de/}}

\maketitle

\begin{abstract}
Process mining is a subfield of process science that analyzes event data collected in databases called event logs. Recently, novel types of event data have become of interest due to the wide industrial application of process mining analyses. In this paper, we examine \emph{uncertain event data}. Such data contain meta-attributes describing the amount of imprecision tied with attributes recorded in an event log. We provide examples of uncertain event data, present the state of the art in regard of uncertainty in process mining, and illustrate open challenges related to this research direction.

\end{abstract}

\section{Introduction}

Process mining is a rapidly growing subfield of data science that aims to automatically analyze event data through a collection of techniques, including the extraction of a process model from a log of historical process executions, the assessment of the conformance and deviations between observed and expected behavior, and the measurement of metrics and indicators over event data and process models. 

The endemic adoption of process mining in the last decades has increased the demand of domain-specific process analysis techniques---for instance, techniques to analyze less traditional types of event data. In this paper, we describe novel types of event data---collectively referred as \emph{uncertain event data}~\cite{DBLP:conf/icpm/PegoraroA19}. Such data contain meta-attributes describing and quantifying the amount of imprecision tied with attributes recorded in an event log. The uncertainty tied to an event attribute might contain indications on its possible values, or also a probability distribution over such values.

The aim of this research direction is to formally illustrate and classify different types of uncertain event data, and develop ad-hoc process mining techniques able to natively function with uncertain event data.

The remainder of the paper is structured as follows. Section~\ref{sec:data} shows examples of uncertain event data. Section~\ref{sec:sources} discusses some possible sources of uncertainty in recorded event data. Section~\ref{sec:related} explores related concepts in process mining and neighboring disciplines. Then, Section~\ref{sec:method} lays out the research methodology and describes the state of the art. Section~\ref{sec:challenges} describes some open challenges in the field of uncertainty in process mining. Finally, Section~\ref{sec:conclusion} concludes the paper.

\section{Uncertainty in Event Data}\label{sec:data}

In order to more clearly visualize the structure of the attributes in uncertain events, let us consider the following process instance, which is a simplified version of actually occurring anomalies, e.g., in the processes of the healthcare domain.

An elderly patient enrolls in a clinical trial for an experimental treatment against myeloproliferative neoplasms, a class of blood cancers. This enrollment includes a lab exam and a visit with a specialist; then, the treatment can begin. The lab exam, performed on the 8th of July, finds a low level of platelets in the blood of the patient, a condition known as thrombocytopenia (TP). During the visit on the 10th of July, the patient reports an episode of night sweats on the night of the 5th of July, prior to the lab exam. The medic notes this but also hypothesizes that it might not be a symptom, since it can be caused either by the condition or by external factors (such as very warm weather). The medic also reads the medical records of the patient and sees that, shortly prior to the lab exam, the patient was undergoing a heparin treatment (a blood-thinning medication) to prevent blood clots. The thrombocytopenia, detected by the lab exam, can then be either primary (caused by the blood cancer) or secondary (caused by other factors, such as a concomitant condition). Finally, the medic finds an enlargement of the spleen in the patient (splenomegaly). It is unclear when this condition has developed: it might have appeared at any moment prior to that point. These events are collected and recorded in the trace shown in Table~\ref{table:uncertaintracestrong} within the hospital's information system.

\begin{table}[t]
	\caption{The \emph{strongly uncertain} trace of an example of healthcare process. The timestamps column shows only the day of the month.}
	\label{table:uncertaintracestrong}
	\centering
	\begin{tabular}{ccccc}
		\textbf{Case ID}        & \textbf{Event ID} & \textbf{Timestamp}                                                                                                     & \textbf{Activity}             & \multicolumn{1}{l}{\textbf{Indeterminacy}} \\ \hline
		\multicolumn{1}{|c|}{ID192} & \multicolumn{1}{c|}{$e_1$} 
		& \multicolumn{1}{c|}{5}                                                                         & \multicolumn{1}{c|}{\emph{NightSweats}}        & \multicolumn{1}{c|}{?}                    \\ \hline
		\multicolumn{1}{|c|}{ID192}& \multicolumn{1}{c|}{$e_2$} & \multicolumn{1}{c|}{8}                                                                         & \multicolumn{1}{c|}{\emph{PrTP}, \emph{SecTP}} & \multicolumn{1}{c|}{}                    \\ \hline
		\multicolumn{1}{|c|}{ID192}& \multicolumn{1}{c|}{$e_3$} & \multicolumn{1}{c|}{4--10}                                                                         & \multicolumn{1}{c|}{\emph{Splenomeg}} & \multicolumn{1}{c|}{}                    \\ \hline
	\end{tabular}
\end{table}

Such scenario, with no known probability, is known as \emph{strong uncertainty}. In this trace, the rightmost column refers to event indeterminacy: in this case, $e_1$ has been recorded, but it might not have occurred in reality, and is marked with a ``?'' symbol. Event $e_2$ has more then one possible activity labels, either \emph{PrTP} or \emph{SecTP}. Lastly, event $e_3$ has an uncertain timestamp, and might have happened at any point in time between the 4th and 10th of July.

Uncertain events may also have probability values associated with them, a scenario defined as \emph{weak uncertainty} (Table~\ref{table:uncertaintraceweak}). In the example described above, suppose the medic estimates that there is a high chance (90\%) that the thrombocytopenia is primary (caused by the cancer). Furthermore, if the splenomegaly is suspected to have developed three days prior to the visit, which takes place on the 10th of July, the timestamp of event $e_3$ may be described through a Gaussian curve with $\mu = 7$. Lastly, the probability that the event $e_1$ has been recorded but did not occur in reality may be known (for example, it may be 25\%).

\begin{table}[t]
	\caption{A trace where uncertain event attributes are labeled with probabilities (\emph{weak uncertainty}).}
	\label{table:uncertaintraceweak}
	\centering
	\begin{tabular}{ccccc}
		\textbf{Case ID}        & \textbf{Event ID} & \textbf{Timestamp}                                                                                                     & \textbf{Activity}             & \multicolumn{1}{l}{\textbf{Indeterminacy}} \\ \hline
		\multicolumn{1}{|c|}{ID348} & \multicolumn{1}{c|}{$e_4$} 
		& \multicolumn{1}{c|}{5}                                                                         & \multicolumn{1}{c|}{\emph{NightSweats}}        & \multicolumn{1}{c|}{$?: 25\%$}                    \\ \hline
		\multicolumn{1}{|c|}{ID348}& \multicolumn{1}{c|}{$e_5$} & \multicolumn{1}{c|}{8}                                                                         & \multicolumn{1}{c|}{\begin{tabular}[c]{@{}c@{}}\emph{PrTP: $90\%$},\\ \emph{SecTP: $10\%$}\end{tabular}} & \multicolumn{1}{c|}{}                    \\ \hline
		\multicolumn{1}{|c|}{ID348}& \multicolumn{1}{c|}{$e_6$} & \multicolumn{1}{c|}{$\mathcal{N}(7, 1)$}                                                                         & \multicolumn{1}{c|}{\emph{Splenomeg}} & \multicolumn{1}{c|}{}                    \\ \hline
	\end{tabular}
\end{table}

Uncertain data as described here can be represented, imported, analyzed and exported on all tools supporting the XES standard~\cite{DBLP:conf/bpm/PegoraroUA21}.

\section{Sources of Uncertainty}\label{sec:sources}

In this section, we will examine some possible sources of uncertain event data. This is not intended to be an exhaustive list nor a proper taxonomy, but it is rather a collection of motivating situations not uncommon in the analysis of event data. In fact, many are documented in literature.

It is important to notice that some causes of uncertainty are \emph{epistemic}, that is, caused by a loss of information or knowledge in some stage of the data recording process; or \emph{aleatoric}, where the uncertainty is intrinsic to the process itself. This distinction, strongly underlined in other fields such as statistics and machine learning, is very important in order to interpret the results of process mining analyses---especially in regard of process improvements prompted by the analysis.

\textbf{Data Coarseness.} Limitations in the precision available to record an event attribute can generate uncertainty. In process mining, this is often the case with timestamps, the attribute we normally rely on to determine a total ordering between events. In some event logs, however, timestamps of different events in the same process trace coincide, because of the coarseness of data recording (e.g., when only the day is recorded but not the time, causing all events happened in the same day to have the same timestamp). This is a source of \emph{partially ordered event data}, a type of uncertain data, and is well documented in process mining research~\cite{DBLP:conf/bpm/LuFA14}.

\textbf{Accuracy of Textual Information.} In many processes, activities and other event attributes are recorded by humans. In such cases, often natural language describes the activity identifier, which may be imprecise in describing what actually happened. For instance, in the uncertain trace of Table~\ref{table:uncertaintracestrong}, the activity label uncertainty of event $e_2$ might have been caused by the activity being recorded simply as ``TP''. This is also a known anomaly in process mining; some approaches to repair it exist, and are based on merging similar labels through NLP methods~\cite{DBLP:journals/is/AaRL21}.

\textbf{Accuracy of Data Detection/Repair Methods.} In some cases, events are not recorded as they happen, but are rather detected from an unstructured source. An example of this is detecting events from video feeds using e.g. deep learning~\cite{DBLP:conf/bpm/CohenG21,DBLP:conf/zeus/LepsienBMK22}. Neural networks are able to predict the occurrence of an event describing it as probability distribution over the possible classes (here, activity labels). This generates probabilistic information about events which fits with the framework described in this paper.

\section{Related Research}\label{sec:related}

Techniques to deal with anomalies and noise in data are present in all branches of data science, from statistics, to machine learning, to process mining itself. Often, a strong focus is on either \emph{filter} anomalous data~\cite{DBLP:journals/access/WangBH19}, and analyze the remaining dataset, or \emph{repair} anomalous attributes, by predicting or inferring heuristically their correct value.

The meta-information describing uncertainty opens a third possibility, which is the development of analysis techniques able to operate on uncertain data as-is. In the context of standard tabular data, this is the research domain of probabilistic databases~\cite{DBLP:series/synthesis/2011Suciu}. Specifically, an approach that lies at the intersection of probabilistic databases and event data analysis is frequent itemsets mining, where the goal is to define frequently-appearing clusters of objects across sets of items (which might be events). There exist approaches to solve this problem for probabilistic data, such as the U-Apriori algorithm~\cite{DBLP:conf/pakdd/ChuiKH07}.

The concept of uncertainty as quantifiable imprecision of data is also of great relevance in the field of machine learning~\cite{DBLP:journals/ml/HullermeierW21}, and very recent research is aimed to detect possible uncertainties in data, quantify them, and classify them as epistemic or aleatoric.

The topic of uncertainty in process mining as defined in this paper is novel, and---to the best of our knowledge---no techniques able to manage uncertainty were described in literature before the start of the doctoral program described in this paper. In the next section, we will describe the research principle that leads our research of uncertain data, and examples of problems solved by process mining techniques applied to uncertain data.

\section{Research Methodology}\label{sec:method}

The premises set out in Sections~\ref{sec:data} and~\ref{sec:sources}, together with the analysis of the literature, brought us to formulate---among others---the following research questions:
\\\textbf{RQ1}: How can we adapt conformance checking to be able to deal with uncertain event data?
\\\textbf{RQ2}: How can we adapt process discovery to be able to deal with uncertain event data?
\\\textbf{RQ3}: How can we embed the mathematical formulation of uncertain event data to obtain uncertain logs from information systems?
\\\textbf{RQ4}: How can we manage the high complexity tied with all possible scenarios described by an uncertain trace?

In the following Subsections~\ref{sec:conformance} and~\ref{sec:discovery} we will describe the methodology utilized to research RQ1 and RQ2, respectively. RQ3 and RQ4 entail challenges that are still completely open, and we comment on them in Section~\ref{sec:challenges}.

Uncertain event data can be considered noise. Filtering or repairing noisy data in a pre-processing step is standard practice both in process mining and data science at large. In our research, the leading principle is the opposite: \emph{retain all data, and exploit the quantification of uncertainty to analyze it in a trustworthy way}. We shift the resolution of uncertainty from the data side to the algorithm side. Such practice avoids information loss and unlocks new insights.

Let us see this principle in action on two of the primary process mining analyses: conformance checking (RQ1) and process discovery (RQ2). 

\subsection{Conformance Checking}\label{sec:conformance}

Conformance checking is one of the main tasks in process mining, and consists in measuring the deviation between process execution data and a reference model. This is particularly useful for organization, since it enables them to compare historical process data against a normative model created by process experts and to identify anomalies in their operations.

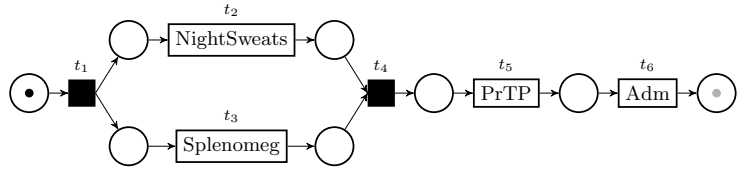
\begin{figure}[t]
	\centering
	\resizebox{.8\textwidth}{!}{%
		\begin{tikzpicture}[node distance=.4cm and .3cm, >=stealth']
		
		\tikzstyle{place} = [circle,draw,thick,minimum size=6mm]
		\tikzstyle{transition} = [rectangle,draw,thick,minimum size=4mm]
		\tikzstyle{invisible} = [transition, fill=black]
		\tikzstyle{finaltoken} = [token, fill=black!30]
		
		\node [place,tokens=1] (p1) {};
		
		\node [invisible] (t0) [right= of p1, label=above:{\scriptsize $t_1$}] {};
		\draw [->] (p1) to (t0.west);
		
		\node [place] (p11) [above right= of t0] {};
		\draw [->] (t0.east) to (p11);
		
		\node [place] (p12) [below right= of t0] {};
		\draw [->] (t0.east) to (p12);
		
		\node [transition] (t1) [right= of p11, label=above:{\scriptsize $t_2$}] {NightSweats};
		\draw [->] (p11) to (t1.west);
		
		\node [place] (p21) [right= of t1] {};
		\draw [->] (t1.east) to (p21);
		
		\node [invisible] (t01) [below right= of p21, label=above:{\scriptsize $t_4$}] {};
		\draw [->] (p21) to (t01.west);
		
		\node [place] (p22) [below left= of t01] {};
		
		\draw [->] (p22) to (t01.west);
		
		\node [transition] (t2) at ($(p12)!0.5!(p22)$) [label=above:{\scriptsize $t_3$}] {Splenomeg};
		\draw [->] (p12) to (t2.west);
		
		\draw [->] (t2.east) to (p22);
		
		\node [place] (p2) [right= of t01] {};
		\draw [->] (t01.east) to (p2);
		
		\node [transition] (t4) [right= of p2, label=above:{\scriptsize $t_5$}] {PrTP};
		\draw [->] (p2) to (t4.west);
		
		\node [place] (p3) [right= of t4] {};
		\draw [->] (t4.east) to (p3);
		
		\node [transition] (t6) [right= of p3, label=above:{\scriptsize $t_6$}] {Adm};
		\draw [->] (p3) to (t6);
		
		\node [place] (p6) [right= of t6] {};
		\draw [->] (t6) to (p6);
		\node [finaltoken] at (p6) {};
		\end{tikzpicture}
	}
	\caption{A normative model for the healthcare process case in the running example. The initial marking is displayed; the gray ``token slot'' represents the final marking.}
	\label{fig:samplemodel}
\end{figure}

Let us assume that we have access to a normative model for the disease of the patient in the running example, shown in Figure~\ref{fig:samplemodel}.

This model essentially states that the disease is characterized by the occurrence of night sweats and splenomegaly on the patient, which may happen concurrently, and then should be followed by primary thrombocytopenia. We would like to measure the conformance between the trace in Table~\ref{table:uncertaintracestrong} and this normative model. A very popular conformance checking technique works via the computation of \emph{alignments}. Through this technique, we are able to identify the deviations in the execution of a process, in the form of behavior happening in the model but not in the trace, and behavior happening in the trace but not in the model. These deviations are identified and used to compute a conformance score between the trace and the process model.

The formulation of alignments in is not applicable to an uncertain trace. In fact, depending on the instantiation of the uncertain attributes of events---like the timestamp of $e_3$ in the trace---the order of event may differ, and so may the conformance score. However, we can look at the best- and worst-case scenarios: the instantiation of attributes of the trace that entails the minimum and maximum number of deviations with respect to the reference model. In our example, two possible outcomes for the sample trace are $\langle \textit{NightSweats}, \textit{Splenomeg}, \textit{PrTP},\allowbreak\textit{Adm} \rangle$ and $\langle \textit{SecTP}, \textit{Splenomeg}, \textit{Adm} \rangle$; both represent the sequence of event that might have happened in reality, but their conformance score is very different. The alignment of the first trace against the reference model can be seen in Table~\ref{table:bestalign}, while the alignment of the second trace can be seen in Table~\ref{table:worstalign}. These two outcomes of the uncertain trace in Table~\ref{table:uncertaintracestrong} represent, respectively, the minimum and maximum amount of deviation possible with respect to the reference model, and define then a lower and upper bound for conformance score.

\begin{table}[]
	\caption{An optimal alignment for $\langle \textit{NightSweats}, \textit{Splenomeg}, \textit{PrTP}, \textit{Adm} \rangle$, one of the possible instantiations of the trace in Table~\ref{table:uncertaintracestrong}, against the model in Figure~\ref{fig:samplemodel}. This alignment has a deviation cost of 0, and corresponds to the best-case scenario for conformance between the process model and the uncertain trace.}
	\label{table:bestalign}
	\centering
	\begin{tabular}{cccccc}
		\multicolumn{1}{|c|}{$\nomove$}	& \multicolumn{1}{c|}{NightSweats}	& \multicolumn{1}{c|}{Splenomeg}	& \multicolumn{1}{c|}{$\nomove$}	& \multicolumn{1}{c|}{PrTP}		& \multicolumn{1}{c|}{Adm}		\\ \hline
		\multicolumn{1}{|c|}{$\tau$}	& \multicolumn{1}{c|}{NightSweats}	& \multicolumn{1}{c|}{Splenomeg}	& \multicolumn{1}{c|}{$\tau$}	& \multicolumn{1}{c|}{PrTP}		& \multicolumn{1}{c|}{Adm}		\\ 
		\multicolumn{1}{|c|}{$t_1$}		& \multicolumn{1}{c|}{$t_2$}	& \multicolumn{1}{c|}{$t_3$}	& \multicolumn{1}{c|}{$t_4$}				& \multicolumn{1}{c|}{$t_5$}	& \multicolumn{1}{c|}{$t_6$}	\\ 
	\end{tabular}
\end{table}

\begin{table}[]
	\caption{An optimal alignment for $\langle \textit{SecTP}, \textit{Splenomeg}, \textit{Adm} \rangle$, one of the possible instantiations of the trace in Table~\ref{table:uncertaintracestrong}, against the model in Figure~\ref{fig:samplemodel}. This alignment has a deviation cost of 3, caused by 2 moves on model and 1 move on log, and corresponds to the worst-case scenario for conformance between the process model and the uncertain trace.}
	\label{table:worstalign}
	\centering
	\begin{tabular}{ccccccc}
		\multicolumn{1}{|c|}{$\nomove$}& \multicolumn{1}{c|}{SecTP}	& \multicolumn{1}{c|}{$\nomove$}	& \multicolumn{1}{c|}{Splenomeg}	& \multicolumn{1}{c|}{$\nomove$}	& \multicolumn{1}{c|}{$\nomove$}				& \multicolumn{1}{c|}{Adm}		\\ \hline
		\multicolumn{1}{|c|}{$\tau$}	& \multicolumn{1}{c|}{$\nomove$}	& \multicolumn{1}{c|}{NightSweats}		& \multicolumn{1}{c|}{Splenomeg}	& \multicolumn{1}{c|}{$\tau$}	& \multicolumn{1}{c|}{PrTP}			& \multicolumn{1}{c|}{Adm}	\\ 
		\multicolumn{1}{|c|}{$t_1$}		& \multicolumn{1}{c|}{}			& \multicolumn{1}{c|}{$t_2$}		& \multicolumn{1}{c|}{$t_3$}	& \multicolumn{1}{c|}{$t_4$}				& \multicolumn{1}{c|}{$t_5$}		& \multicolumn{1}{c|}{$t_6$}		\\ 
	\end{tabular}
\end{table}

\begin{figure}[t]
	\centering
	\resizebox{.8\textwidth}{!}{%
		\begin{tikzpicture}[node distance=.6cm and .9cm, >=stealth']
		
		\tikzstyle{place} = [circle,draw,thick,minimum size=6mm]
		\tikzstyle{transition} = [rectangle,draw,thick,minimum size=4mm]
		\tikzstyle{invisible} = [transition, fill=black]
		\tikzstyle{finaltoken} = [token, fill=black!30]
		
		\node [place,tokens=1] (p1) [label=above:{\scriptsize $(\textsc{start}, e_1)$}] {};
		
		\node [transition] (t1) [above right= of p1, label=above:{\scriptsize $(e_1, NightSweats)$}] {NightSweats};
		\draw [->] (p1) to (t1.west);
		
		\node [invisible] (t2) [below right= of p1, label=above:{\scriptsize $(e_1, \tau)$}] {NightSweats};
		\draw [->] (p1) to (t2.west);
		
		\node [place] (p2) [below right= of t1, label=above:{\scriptsize $(e_1, e_2)$}] {};
		\draw [->] (t1.east) to (p2);
		\draw [->] (t2.east) to (p2);
		
		\node [transition] (t3) [above right= of p2, label=above:{\scriptsize $(e_2, PrTP)$}] {PrTP};
		\draw [->] (p2) to (t3.west);
		
		\node [transition] (t4) [below right= of p2, label=above:{\scriptsize $(e_2, SecTP)$}] {SecTP};
		\draw [->] (p2) to (t4.west);
		
		\node [place] (p3) [below right= of t3, label=above:{\scriptsize $(e_2, e_4)$}] {};
		\draw [->] (t3.east) to (p3);
		\draw [->] (t4.east) to (p3);
		
		\node [place,tokens=1] (p4) [below left= of t2, label=above:{\scriptsize $(\textsc{start}, e_3)$}] {};
		
		\node [place] (p5) [below right= of t4, label=above:{\scriptsize $(e_3, e_4)$}] {};
		
		\node [transition] (t5) at ($(p4)!0.5!(p5)$) [label=above:{\scriptsize $(e_3, Splenomeg)$}] {Splenomeg};
		\draw [->] (p4) to (t5);
		\draw [->] (t5) to (p5);
		
		\node [transition] (t6) [above right= of p5, label=above:{\scriptsize $(e_4, Adm)$}] {Adm};
		\draw [->] (p3) to (t6.north west);
		\draw [->] (p5) to (t6.south west);
		
		\node [place] (p6) [right= of t6, label=above:{\scriptsize $(e_4, \textsc{end})$}] {};
		\draw [->] (t6) to (p6);
		\node [finaltoken] at (p6) {};
		\end{tikzpicture}
	}
	\caption{The behavior net~\cite{DBLP:conf/bis/PegoraroUA20} representing the behavior of the uncertain trace in Table~\ref{table:uncertaintracestrong}. The initial marking is displayed; the gray ``token slot'' represents the final marking. This artifact is necessary to perform conformance checking between uncertain traces and a reference model.}
	\label{fig:bn}
\end{figure}
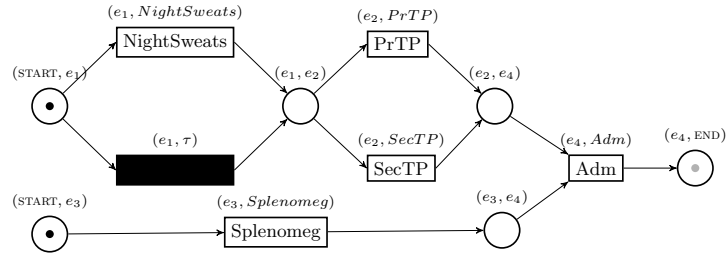

It is possible to find bounds for the conformance score of an uncertain trace and a reference process model with an extension of the alignment technique~\cite{DBLP:journals/is/PegoraroUA21}. In order to find such bounds, it is necessary to build a Petri net able to simulate all possible behaviors in the uncertain trace, called the \emph{behavior net}~\cite{DBLP:journals/algorithms/PegoraroUA20}. The behavior net of the trace in Table~\ref{table:uncertaintracestrong} is shown in Figure~\ref{fig:bn}.

The alignments in Tables~\ref{table:bestalign} and~\ref{table:worstalign} show how we can get actionable insights from process mining over uncertain data. In some applications it is reasonable and appropriate to remove uncertain data from an event log via filtering, and then compute log-level aggregate information---such as total number of deviations, or average deviations per trace---using the remaining certain data. Even in processes where this is possible, doing so prevents the important process mining task of case diagnostic. Conversely, uncertain alignments allow not only to have best- and worst-case scenarios for a trace, but also to individuate the specific deviations affecting both scenarios. For instance, the alignments of the running example can be implemented in a system that warns the medics that the patient might have been affected by a secondary thrombocytopenia not explained by the model of the disease. Since the model indicates that the disease should develop primary thrombocytopenia as a symptom, this patient is at risk of both types of platelets deficit simultaneously, which is a serious condition. The medics can then intervene to avoid this complication, and perform more exams to ascertain the cause of the patient's thrombocytopenia.

\subsection{Process Discovery}\label{sec:discovery}

Process discovery is another main objective in process mining, and involves automatically creating a process model from event data. Many process discovery algorithms rely on the concept of \emph{directly-follows relationships} between activities to gather clues on how to structure the process model. \emph{Uncertain Directly-Follows Graphs} (UDFGs) enable the representation of directly-follows relationships in an uncertain event log; they consist in directed graphs where the activity labels appearing in the event log constitute the nodes, and the edges are decorated with information on the minimum and maximum frequency observable for the directly-follows relation between pair of activities.

Let us examine an example of UDFG. In order to build a significant example, we need to introduce an entire uncertain event log; since the full table notation for uncertain traces becomes cumbersome for entire logs, let us utilize a shorthand simplified notation. In a trace, we represent an uncertain event with multiple possible activity labels by listing all the associated labels between curly braces.

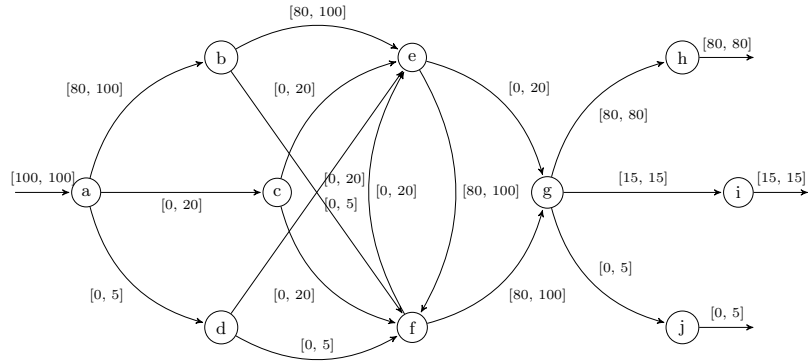
\begin{figure}[t]
	\centering
	\resizebox{.9\textwidth}{!}{%
		\begin{tikzpicture}[->,>=stealth',shorten >=1pt,node distance=3.4cm,auto,main node/.style={circle,draw,align=center}]
		\node[main node]	(A)	[]						{a};
		\node[main node]	(B)	[above right of=A]		{b};
		\node[main node]	(C)	[right of=A]			{c};
		\node[main node]	(D)	[below right of=A]		{d};
		\node[main node]	(E)	[above right of=C]		{e};
		\node[main node]	(F)	[below right of=C]		{f};
		\node[main node]	(G)	[below right of=E]		{g};
		\node[main node]	(H)	[above right of=G]		{h};
		\node[main node]	(I)	[right of=G]			{i};
		\node[main node]	(J)	[below right of=G]		{j};
		\node[left=1cm of A] (K) {};
		\node[right=1cm of H] (L) {};
		\node[right=1cm of I] (M) {};
		\node[right=1cm of J] (N) {};
		\path
		(A) edge [bend left] node {\scriptsize [80, 100]} (B)
		(A) edge node [swap] {\scriptsize [0, 20]} (C)
		(A) edge [bend right] node [swap] {\scriptsize [0, 5]} (D)
		(B) edge [bend left] node {\scriptsize [80, 100]} (E)
		(B) edge node {\scriptsize [0, 20]} (F)
		(C) edge [bend left] node {\scriptsize [0, 20]} (E)
		(C) edge [bend right] node [swap] {\scriptsize [0, 20]} (F)
		(D) edge node [swap] {\scriptsize [0, 5]} (E)
		(D) edge [bend right] node {\scriptsize [0, 5]} (F)
		(E) edge [bend left] node {\scriptsize [80, 100]} (F)
		(F) edge [bend left] node [swap] {\scriptsize [0, 20]} (E)
		(E) edge [bend left] node {\scriptsize [0, 20]} (G)
		(F) edge [bend right] node [swap] {\scriptsize [80, 100]} (G)
		(G) edge [bend left] node [swap] {\scriptsize [80, 80]} (H)
		(G) edge node {\scriptsize [15, 15]} (I)
		(G) edge [bend right] node {\scriptsize [0, 5]} (J)
		(K) edge node {\scriptsize [100, 100]} (A)
		(H) edge node {\scriptsize [80, 80]} (L)
		(I) edge node {\scriptsize [15, 15]} (M)
		(J) edge node {\scriptsize [0, 5]} (N)
		;
		\end{tikzpicture}
	}
	\caption{The \emph{Uncertain Directly-Follows Graph} (UDFG) computed based on the uncertain event log $\langle a, b, e, f, g, h \rangle^{80}$, $\langle a, \{b, c\}, [e, f], g, i \rangle^{15}$, $\langle a, \{b, c, d\}, [e, f], g, \overline{j} \rangle^{5}$. The arcs are labeled with the minimum and maximum number of directly-follows relationship observable between activities in the corresponding trace. The construction of this object is necessary to perform automatic process discovery over uncertain event data.}
	\label{fig:udfg}
\end{figure}

When two events have mutually overlapping timestamps, we write their activity labels between square brackets, and we indicate indeterminate events by overlining them. For instance, the trace $\langle \overline{a}, \{b, c\}, [d, e] \rangle$ is a trace containing 4 events, of which the first is an indeterminate event with activity label $a$, the second is an uncertain event that can have either $b$ or $c$ as activity label, and the last two events have an interval as timestamp (and the two ranges overlap). Let us consider the following event log:

$\langle a, b, e, f, g, h \rangle^{80}, \langle a, \{b, c\}, [e, f], g, i \rangle^{15}, \langle a, \{b, c, d\}, [e, f], g, \overline{j} \rangle^{5}$.

For each pair of activities, we can count the minimum and maximum occurrences of a directly-follows relationship that can be observed in the log. The resulting UDFG is shown in Figure~\ref{fig:udfg}.

\clearpage

\begin{figure}[h]
	\centering
	\subcaptionbox{A process model that can only replay the relationships appearing in the certain parts of the traces in the uncertain log. Here, information from uncertainty has been excluded completely.\label{subfig:1}\\}{%
		\resizebox{.8\textwidth}{!}{%
			\begin{tikzpicture}[node distance=.1cm and .3cm, >=stealth']
			
			\tikzstyle{place} = [circle,draw,thick,minimum size=6mm]
			\tikzstyle{transition} = [rectangle,draw,thick,minimum size=4mm]
			\tikzstyle{invisible} = [transition, fill=black]
			\tikzstyle{finaltoken} = [token, fill=black!30]
			
			\node [place,tokens=1] (p1) {};
			\node [transition] (A) [right= of p1] {a};
			\node [place] (p2) [right= of A] {};
			\node [transition] (B) [right= of p2] {b};
			\node [place] (p3) [right= of B] {};
			\node [transition] (E) [right= of p3] {e};
			\node [place] (p4) [right= of E] {};
			\node [transition] (F) [right= of p4] {f};
			\node [place] (p5) [right= of F] {};
			\node [transition] (G) [right= of p5] {g};
			\node [place] (p6) [right= of G] {};
			\node [transition] (H) [above right= of p6] {h};
			\node [transition] (I) [below right= of p6] {i};
			\node [place] (p7) [below right= of H] {};
			\node [finaltoken] at (p7) {};
			
			\draw [->] (p1) to (A.west);
			\draw [->] (A.east) to (p2);
			\draw [->] (p2) to (B.west);
			\draw [->] (B.east) to (p3);
			\draw [->] (p3) to (E.west);
			\draw [->] (E.east) to (p4);
			\draw [->] (p4) to (F.west);
			\draw [->] (F.east) to (p5);
			\draw [->] (p5) to (G.west);
			\draw [->] (G.east) to (p6);
			\draw [->] (p6) to (H.west);
			\draw [->] (p6) to (I.west);
			\draw [->] (H.east) to (p7);
			\draw [->] (I.east) to (p7);
			\end{tikzpicture}
		}
	}%
	\hfill
	\subcaptionbox{A process model that can replay some---but not all---the relationships appearing in the uncertain parts of the traces in the uncertain log. This process model mediates between representing only certain observation and representing all the possible behavior in the process.\label{subfig:2}\\}{%
		\resizebox{.8\textwidth}{!}{%
			\begin{tikzpicture}[node distance=.1cm and .3cm, >=stealth']
			
			\tikzstyle{place} = [circle,draw,thick,minimum size=6mm]
			\tikzstyle{transition} = [rectangle,draw,thick,minimum size=4mm]
			\tikzstyle{invisible} = [transition, fill=black]
			\tikzstyle{finaltoken} = [token, fill=black!30]
			
			\node [place,tokens=1] (p1) {};
			\node [transition] (A) [right= of p1] {a};
			\node [place] (p2) [right= of A] {};
			\node [transition] (B) [above right= of p2] {b};
			\node [transition] (C) [below right= of p2] {c};
			\node [place] (p22) [below right= of B] {};
			\node [invisible] (Z) [right= of p22] {};
			\node [place] (p3) [above right= of Z] {};
			\node [place] (p31) [below right= of Z] {};
			\node [transition] (E) [right= of p3] {e};
			\node [transition] (F) [right= of p31] {f};
			\node [place] (p51) [right= of E] {};
			\node [place] (p5) [right= of F] {};
			\node [transition] (G) [above right= of p5] {g};
			\node [place] (p6) [right= of G] {};
			\node [transition] (H) [above right= of p6] {h};
			\node [transition] (I) [below right= of p6] {i};
			\node [place] (p7) [below right= of H] {};
			\node [finaltoken] at (p7) {};
			
			\draw [->] (p1) to (A.west);
			\draw [->] (A.east) to (p2);
			\draw [->] (p2) to (B.west);
			\draw [->] (p2) to (C.west);
			\draw [->] (B.east) to (p22);
			\draw [->] (C.east) to (p22);
			\draw [->] (p22) to (Z.west);
			\draw [->] (Z.east) to (p3);
			\draw [->] (Z.east) to (p31);
			\draw [->] (p3) to (E.west);
			\draw [->] (p31) to (F.west);
			\draw [->] (E.east) to (p51);
			\draw [->] (F.east) to (p5);
			\draw [->] (p51) to (G.west);
			\draw [->] (p5) to (G.west);
			\draw [->] (G.east) to (p6);
			\draw [->] (p6) to (H.west);
			\draw [->] (p6) to (I.west);
			\draw [->] (H.east) to (p7);
			\draw [->] (I.east) to (p7);
			\end{tikzpicture}
		}
	}%
	\hfill
	\subcaptionbox{A process model that can replay all possible configurations of certain and uncertain traces in the uncertain log. This process model has the highest possible replay fitness, but is also very likely to contain some noisy or otherwise unwanted behavior.\label{subfig:3}\\}{%
		\resizebox{.8\textwidth}{!}{%
			\begin{tikzpicture}[node distance=.1cm and .3cm, >=stealth']
			
			\tikzstyle{place} = [circle,draw,thick,minimum size=6mm]
			\tikzstyle{transition} = [rectangle,draw,thick,minimum size=4mm]
			\tikzstyle{invisible} = [transition, fill=black]
			\tikzstyle{finaltoken} = [token, fill=black!30]
			
			\node [place,tokens=1] (p1) {};
			\node [transition] (A) [right= of p1] {a};
			\node [place] (p2) [right= of A] {};
			\node [transition] (B) [above right= of p2] {b};
			\node [transition] (D) [ below right= of p2] {d};
			\node [place] (p22) [below right= of B] {};
			\node [transition] (C)  at ($(p2)!0.5!(p22)$) {c};
			\node [invisible] (Z) [right= of p22] {};
			\node [place] (p3) [above right= of Z] {};
			\node [place] (p31) [below right= of Z] {};
			\node [transition] (E) [right= of p3] {e};
			\node [transition] (F) [right= of p31] {f};
			\node [place] (p51) [right= of E] {};
			\node [place] (p5) [right= of F] {};
			\node [transition] (G) [above right= of p5] {g};
			\node [place] (p6) [right= of G] {};
			
			\node [transition] (I) [above right= of p6] {i};
			\node [transition] (J) [below right= of p6] {j};
			\node [transition] (H) [above= of I] {h};
			\node [invisible] (K) [below= of J] {k};
			\node [place] (p7) [below right= of I] {};
			\node [finaltoken] at (p7) {};
			
			\draw [->] (p1) to (A.west);
			\draw [->] (A.east) to (p2);
			\draw [->] (p2) to (B.west);
			\draw [->] (p2) to (C.west);
			\draw [->] (p2) to (D.west);
			\draw [->] (B.east) to (p22);
			\draw [->] (C.east) to (p22);
			\draw [->] (D.east) to (p22);
			\draw [->] (p22) to (Z.west);
			\draw [->] (Z.east) to (p3);
			\draw [->] (Z.east) to (p31);
			\draw [->] (p3) to (E.west);
			\draw [->] (p31) to (F.west);
			\draw [->] (E.east) to (p51);
			\draw [->] (F.east) to (p5);
			\draw [->] (p51) to (G.west);
			\draw [->] (p5) to (G.west);
			\draw [->] (G.east) to (p6);
			\draw [->] (p6) to (H.west);
			\draw [->] (p6) to (I.west);
			\draw [->] (p6) to (J.west);
			\draw [->] (p6) to (K.west);
			\draw [->] (H.east) to (p7);
			\draw [->] (I.east) to (p7);
			\draw [->] (J.east) to (p7);
			\draw [->] (K.east) to (p7);
			\end{tikzpicture}
		}
	}%
	\caption{Three different process models for the uncertain event log $\langle a, b, e, f, g, h \rangle^{80}$, $\langle a, \{b, c\}, [e, f], g, i \rangle^{15}$, $\langle a, \{b, c, d\}, [e, f], g, \overline{j} \rangle^{5}$ obtained through inductive mining over an uncertain directly-follows graph. The different filtering parameters for the UDFG yield models with distinct features.}
	\label{fig:uncertaindiscovery}
\end{figure}

This graph can be then utilized to discover process models of uncertain logs via process discovery methods based on directly-follows relationships. In a previous work we illustrated this principle by applying it to the inductive miner, a popular discovery algorithm~\cite{DBLP:conf/bpm/PegoraroUA19}; the edges of the UDFG can be filtered using the information on the labels, in such a way that the final model can represent all possible behavior in the uncertain log, or only a part. Figure~\ref{fig:uncertaindiscovery} shows some process models obtained through inductive mining of the UDFG, as well as a description regarding how the model relates to the original uncertain log. Notice how all three models in the figure are not obtainable by filtering out the traces with uncertainty from the log; this would radically remove useful information from the event log.

The process mining techniques described here are available in a Python library built on the PM4Py framework~\cite{DBLP:conf/apn/PegoraroUA21}.

\section{Open Challenges}\label{sec:challenges}

The examples shown in the previous section show some viable solutions to typical process mining problems in the uncertain case; however, many technical challenges remain open.

A prominent problem is in \emph{data sourcing} (RQ3). At the present time, no information system natively supports the quantification of uncertainty, thus examples of uncertain logs come from pre-processing steps that label data as uncertain based on domain knowledge provided by process experts. This needs to be automated; for instance, intervening directly on the process of data recording. \emph{Uncertainty-aware information systems} would not only enable the full automation of techniques for process mining over uncertainty, but also more reliably support general data mining techniques, which would gain an additional measure of reliability.

Retaining all information from uncertain traces has the problem that the possible behavior are subject to a \emph{combinatorial explosion} (RQ4). While techniques to fully describe all behavior and related probabilities exists~\cite{DBLP:conf/icpm/0001BUA21}, this comes at the cost of high (sometimes exponential) computational complexity. In existing techniques, this has been mitigated by representing uncertain traces as graphs (e.g., the behavior net), and designing algorithms able to work on graphs as inputs. However, this is ineffective for some applications, such as measuring classic model/log metrics in process mining like fitness and precision. We might overcome this problem by switching to approximated techniques, which allow to trade-off speed and accuracy in a controlled manner.

\section{Conclusion}\label{sec:conclusion}

The research field of process mining on uncertain event data, while at its infancy, has proven useful in solving real-life problems that can appear on uncertain data and that require dedicated techniques. Such techniques do not filter out or repair the uncertain attributes in event logs, but rather use extended versions of known process mining algorithms to obtain an uncertainty-aware solution---a solution that explains uncertainty as intrinsic part of the process. 

In pursuing this line of research, we aim to create a comprehensive set of techniques that allow to carry out the most typical process mining tasks on data with quantified uncertainty. Our future work will be guided by the open challenges hereby described which, once solved, will enable a rich array of analysis techniques on uncertain data.

\bibliographystyle{splncs04}
\bibliography{bibliography}

\end{document}